\documentclass[11pt,a4paper]{article}

\usepackage[utf8]{inputenc}
\usepackage[T1]{fontenc}
\usepackage{lmodern}
\usepackage[margin=1in]{geometry}
\usepackage{amsmath,amssymb,amsthm}
\usepackage{graphicx}
\usepackage{booktabs}
\usepackage{hyperref}
\usepackage{cleveref}
\usepackage{natbib}
\usepackage{enumitem}
\usepackage{xcolor}
\usepackage{microtype}
\usepackage{orcidlink}
\usepackage{tabularx}
\usepackage{listings}
\usepackage{tikz}
\usetikzlibrary{arrows.meta, positioning, calc, fit, backgrounds}
\usepackage[most]{tcolorbox}

\hypersetup{
  colorlinks=true,
  linkcolor=blue!70!black,
  citecolor=green!50!black,
  urlcolor=blue!60!black,
}

\definecolor{accentblue}{HTML}{4A7FB5}

\newtheorem{definition}{Definition}
\newtheorem{thesis}{Thesis}
\newtheorem{conjecture}{Conjecture}

\tcolorboxenvironment{definition}{
  blanker, borderline west={2.5pt}{0pt}{accentblue},
  left=8pt, top=4pt, bottom=4pt,
  before skip=10pt, after skip=10pt,
  breakable,
}
\tcolorboxenvironment{thesis}{
  blanker, borderline west={2.5pt}{0pt}{accentblue},
  left=8pt, top=4pt, bottom=4pt,
  before skip=10pt, after skip=10pt,
  breakable,
}
\tcolorboxenvironment{conjecture}{
  blanker, borderline west={2.5pt}{0pt}{accentblue},
  left=8pt, top=4pt, bottom=4pt,
  before skip=10pt, after skip=10pt,
  breakable,
}

\title{%
  \textbf{Context Cartography: Toward Structured Governance of Contextual Space in Large Language Model Systems}%
}

\author{
  Zihua Wu\,\orcidlink{0000-0002-4903-4485} \\
  NVIDIA \\
  \texttt{zihuaw@nvidia.com}
  \and
  Georg Gartner\,\orcidlink{0000-0003-2002-5339} \\
  Research Division Cartography, TU Wien \\
  \texttt{georg.gartner@tuwien.ac.at}
}

\date{March 2026}

\begin{document}
\maketitle

\begin{abstract}
The prevailing approach to improving large language model (LLM)
reasoning has centered on expanding context windows, implicitly
assuming that more tokens yield better performance. However,
empirical evidence---including the ``lost in the middle'' effect and
long-distance relational degradation---demonstrates that contextual
space exhibits structural gradients, salience asymmetries, and entropy
accumulation under transformer architectures. We introduce
\textbf{Context Cartography}, a formal framework for the deliberate
governance of contextual space. We define a tripartite zonal model
partitioning the informational universe into black fog (unobserved),
gray fog (stored memory), and the visible field (active reasoning
surface), and formalize seven cartographic operators---reconnaissance,
selection, simplification, aggregation, projection, displacement, and
layering---as transformations governing information transitions between
and within zones. The operators are derived from a systematic coverage
analysis of all non-trivial zone transformations and are organized by
transformation type (what the operator does) and zone scope (where it
applies). We ground the framework in the salience geometry of
transformer attention, characterizing cartographic operators as
necessary compensations for linear prefix memory, append-only state,
and entropy accumulation under expanding context. An analysis of four contemporary
systems (Claude Code, Letta, MemOS, and OpenViking) provides
interpretive evidence that these operators are converging independently
across the industry.
We derive testable predictions from the framework---including
operator-specific ablation hypotheses---and propose a diagnostic
benchmark for empirical validation.
\end{abstract}

\medskip
\noindent\textbf{Keywords:} context engineering, large language models,
agent architecture, memory systems, cartographic generalization,
attention mechanisms

\section{Introduction}
\label{sec:introduction}

The dominant approach to improving LLM reasoning over the past several
years has been to expand context windows---from a few thousand tokens to
over one million~\citep{ding2024survey}. This expansion widens the
model's field of view but does not govern how information moves between
sensing, memory, and reasoning; without such governance, visibility
becomes overload. Models exhibit severe performance degradation when
critical information occupies intermediate
positions~\citep{liu2024lost}, relational knowledge degrades as token
distance increases~\citep{li2025lost}, and effective context utilization
falls dramatically short of stated capacity~\citep{paulsen2025mecw}.

The emerging discipline of Context Engineering~\citep{mei2025cesurvey}
addresses this challenge broadly, encompassing context retrieval,
processing, and management. However, it lacks a formal spatial model of
\emph{how} contextual space is structured and why certain management
strategies are architecturally necessary. Cartography, understood
broadly, is the study of the structured representation of large
information spaces under bounded
cognition~\citep{maceachren1995maps, robinson1976nature}. While
historically grounded in geographic map-making, cartographic
principles---selection, generalization, controlled distortion,
scale-dependent representation---apply wherever a bounded medium must
faithfully represent an unbounded
reality~\citep{gartner2007ubiquitous}: cache hierarchies decide which
objects to retain under capacity limits, and information visualization
projects high-dimensional data onto two-dimensional displays under
perceptual constraints. In each case, a bounded medium must represent
an unbounded source through controlled selection, compression, and
distortion. We introduce \textbf{Context Cartography}---a
framework that extends cartographic generalization
theory~\citep{mcmaster1992generalization} to the governance of
contextual space in LLM systems, treating LLM context as the latest instance of this general
bounded-representation problem---not a transcript to be extended but
\emph{terrain to be mapped}, with boundaries, gradients, and
transitions that require deliberate governance.

The analogy to the ``fog of war''---a term originating in
Clausewitz's theory of military uncertainty~\citep{clausewitz1832war}
and later operationalized in real-time strategy games---is
instructive. In such games, uncertainty is not binary but stratified:
\emph{black fog} (unexplored territory), \emph{gray fog} (previously
revealed but no longer visible), and the \emph{visible field}
(currently illuminated). Skilled players manage transitions between
these zones rather than attempting to illuminate everything
simultaneously. LLM-based agents face the same structure of
uncertainty across context.

Our central claim is that context is not a passive container but a
structured spatial field whose governance is architecturally necessary.
The specific salience geometry we characterize is grounded in
transformer attention, but the zonal model and operators are
architecture-general. Sequential architectures exhibit positional
sensitivity (transformers via attention
non-uniformity~\citep{liu2024lost}; state space models and linear
attention architectures via recency-dominant state
compression~\citep{gu2024mamba, peng2025rwkv7}); non-sequential
architectures such as diffusion language
models~\citep{nie2025llada} face different constraints (computational
cost scaling with sequence length) that shift operator priorities
without eliminating the need for governance. The zonal structure itself persists
regardless of architecture: the world exceeds any model's capacity,
memory requires structure, and reasoning requires a bounded surface.

The contributions of this paper are fourfold:
\begin{enumerate}[leftmargin=*]
  \item We formalize a tripartite zonal model of contextual space with
        explicit definitions, state transitions, and failure modes
        (\Cref{sec:zones}).
  \item We characterize the salience geometry of transformer context
        windows, establishing why cartographic governance is
        architecturally necessary (\Cref{sec:geometry}).
  \item We define seven cartographic operators---derived from a
        systematic coverage analysis of all zone transformations---as
        formal transformations organized by transformation type and
        zone scope, with specified cartographic correspondences
        (\Cref{sec:operators}).
  \item We derive testable predictions from the framework, propose a
        diagnostic benchmark, and outline a research agenda for
        empirical validation (\Cref{sec:agenda}).
\end{enumerate}

\section{Related Work}
\label{sec:related}

\paragraph{Long-Context Language Models.}
Techniques including sparse attention~\citep{beltagy2020longformer},
hardware-aware attention implementation~\citep{dao2022flashattention}, rotary
positional encodings, and KV cache optimization~\citep{li2025kvcache}
have enabled context lengths exceeding one million
tokens~\citep{ding2024survey}. However, \citet{liu2024lost}
demonstrated that performance degrades sharply when relevant information
occupies intermediate positions. Subsequent work has established that
this positional bias emerges from the structural properties of causal
attention~\citep{wu2025emergence}. Empirical benchmarks confirm
the effect across model families~\citep{gupte2025gm}. Mitigations such as
multi-scale positional encoding~\citep{he2024position}, hidden-state
scaling~\citep{yu2025mitigate}, and attention
calibration~\citep{hsieh2024found} offer partial remedies but do not
address the fundamental geometry of contextual space. Similarly,
hierarchical and sparse attention patterns (e.g.,
Longformer~\citep{beltagy2020longformer}) and recurrent memory
architectures (e.g., RWKV~\citep{peng2025rwkv7}) reshape the salience
function but do not govern how information moves between zones;
Context Cartography operates at the governance layer above
architectural attention patterns.

\paragraph{Prompt Engineering and Chain-of-Thought.}
Chain-of-Thought prompting~\citep{wei2022cot} demonstrated that
structuring the reasoning trace within the context window significantly
improves performance on complex tasks. This can be understood as an
implicit application of the layering operator: separating the reasoning
trace from the task specification within the same sequence. More broadly,
prompt engineering operates at the level of local phrasing, while
Context Cartography operates at the level of spatial organization across
the entire context.

\paragraph{Retrieval-Augmented Generation.}
RAG~\citep{lewis2020rag} brings information from the black fog into the
visible field through retrieval. Surveys by~\citet{gao2024rag} catalog
the evolution of retrieval strategies. However, retrieval from a pre-indexed corpus operates primarily as a
selection and projection mechanism---determining \emph{what} to
recall from stored memory without systematically addressing
\emph{how} recalled content should be positioned or layered within the
context window~\citep{cuconasu2024power, shi2023large}.

\paragraph{Memory-Augmented Agent Architectures.}
Tool-augmented agents~\citep{schick2023toolformer, yao2023react} extend
LLM capabilities through external function calls.
MemGPT~\citep{packer2024memgpt} models context management as an
operating system problem. MemOS~\citep{memos2025} introduces a memory operating system with
structured, typed memory containers (MemCubes). Generative Agents~\citep{park2023generative} maintain persistent
memory streams that agents retrieve and reflect upon. Memorizing
Transformers~\citep{wu2022memorizing} augment attention with retrieval
over external memory stores. Recent surveys~\citep{liu2025memorysurvey}
identify three dominant memory realizations (token-level, parametric,
latent) with distinct functional roles. These systems implicitly adopt
principles we formalize as cartographic operators.

\paragraph{Context Engineering.}
\citet{mei2025cesurvey} establish Context Engineering as a broad
discipline encompassing retrieval, processing, and management of
context for LLMs. Context Cartography is complementary: where Context
Engineering provides a comprehensive taxonomy of \emph{what} techniques
exist, Context Cartography provides a formal spatial model of
\emph{why} certain techniques are necessary and \emph{how} they relate
to the geometry of contextual space. Specifically, Context Cartography
adds: (i)~a zonal model partitioning the informational universe into
epistemic states, (ii)~a geometric analysis grounding management
strategies in attention physics, and (iii)~an operator formalism
derived from cartographic generalization theory---enabling
distinctions (e.g., archival vs.\ destructive summarization) that a
flat taxonomy does not capture.

\paragraph{Cartographic Generalization and Ubiquitous Cartography.}
Cartographic generalization reduces map information content while
preserving essential
characteristics~\citep{mcmaster1992generalization}. Standard operators
include selection, simplification, aggregation, typification,
displacement, collapse, and
symbolization~\citep{roth2011cartographic, stoter2014state}.
\citet{brassel1988generalization} established the canonical process
model for automated generalization---structure recognition, process
modeling, execution---that our cartographic pipeline concept
(\Cref{sec:operators}) extends to contextual space.
\citet{robinson1976nature} formalized cartographic communication as an
encoding--transmission--decoding channel, and
\citet{maceachren1995maps} integrated cognitive and semiotic
approaches to show that maps function as abstract synthetic
representations under bounded cognition.
\citet{gartner2007ubiquitous} argue that cartographic principles
extend beyond geographic maps to any context where spatial information
is communicated under constraints---a position we adopt and extend to
LLM context management. Recent work has noted that spatial hierarchies
from cartography improve LLM performance on entity disambiguation
tasks~\citep{gartner2025cartography}, and the adaptive cartographic
methodologies surveyed by~\citet{lin2011cartography} for virtual
geographic environments anticipate patterns now emerging in agent
architecture design.

\section{The Context Cartography Framework}
\label{sec:framework}

\subsection{Contextual Space and Epistemic Zones}
\label{sec:zones}

\begin{definition}[Contextual Universe]
\label{def:universe}
Let $\mathcal{U}$ denote the \emph{contextual universe}---the totality
of information potentially relevant to an agent's task. At any time
step $t$, $\mathcal{U}$ is partitioned into three disjoint epistemic
zones:
\[
  \mathcal{U} = \mathcal{B}_t \;\cup\; \mathcal{G}_t \;\cup\; \mathcal{V}_t
\]
where $\mathcal{B}_t$ (black fog) is the unobserved frontier,
$\mathcal{G}_t$ (gray fog) is stored memory, and $\mathcal{V}_t$
(visible field) is the active reasoning surface.
\end{definition}

\begin{definition}[Zone Transitions]
\label{def:transitions}
Information moves between zones via four transition functions:
\begin{align}
  \textup{\textsc{sense}}   &: \mathcal{B} \to \mathcal{G}
    && \text{(tool calls, search, code execution)} \\
  \textup{\textsc{recall}} &: \mathcal{G} \to \mathcal{V}
    && \text{(retrieval, memory loading)} \\
  \textup{\textsc{evict}}   &: \mathcal{V} \to \mathcal{G}
    && \text{(summarization, archival)} \\
  \textup{\textsc{expire}}  &: \mathcal{G} \to \mathcal{B}
    && \text{(staleness, invalidation)}
\end{align}
\end{definition}

\begin{figure}[t]
\centering
\begin{tikzpicture}[
  zone/.style={draw, rounded corners=6pt, minimum width=3.2cm,
    minimum height=2.0cm, align=center, font=\small},
  trans/.style={-{Stealth[length=5pt]}, thick, font=\scriptsize},
  fail/.style={font=\scriptsize\itshape, text=red!70!black},
  >=Stealth
]
\node[zone, fill=black!8] (B) at (0, 0)
  {$\mathcal{B}$\\[2pt]\scriptsize Black Fog\\[-1pt]\scriptsize (Unobserved)};
\node[zone, fill=black!4] (G) at (4.8, 0)
  {$\mathcal{G}$\\[2pt]\scriptsize Gray Fog\\[-1pt]\scriptsize (Stored Memory)};
\node[zone, fill=white] (V) at (9.6, 0)
  {$\mathcal{V}$\\[2pt]\scriptsize Visible Field\\[-1pt]\scriptsize (Active Context)};

\draw[trans] ([yshift=4pt]B.east) -- node[above, yshift=1pt]
  {\textsc{sense}} ([yshift=4pt]G.west);
\draw[trans] ([yshift=4pt]G.east) -- node[above, yshift=1pt]
  {\textsc{recall}} ([yshift=4pt]V.west);
\draw[trans] ([yshift=-4pt]V.west) -- node[below, yshift=-1pt]
  {\textsc{evict}} ([yshift=-4pt]G.east);
\draw[trans] ([yshift=-4pt]G.west) -- node[below, yshift=-1pt]
  {\textsc{expire}} ([yshift=-4pt]B.east);

\draw[trans, densely dashed, gray]
  ([yshift=12pt]B.east) to[bend left=28]
  node[above, yshift=1pt, text=gray]
  {\textsc{recall}$\,\circ\,$\textsc{sense}}
  ([yshift=12pt]V.west);

\node[fail, below=6pt of B] {Hallucination};
\node[fail, below=6pt of G] {Drift, bloat};
\node[fail, below=6pt of V] {Overload, dilution};
\end{tikzpicture}
\caption{The tripartite zonal architecture.  Solid arrows show the
four named transitions; the dashed arc shows the degenerate
$\mathcal{B} \to \mathcal{V}$ path that Context Cartography
recommends mediating through $\mathcal{G}$ or operator guards
($\phi$, $\pi^{+}$).  Zone-level failure modes appear below each
zone.}
\label{fig:zones}
\end{figure}
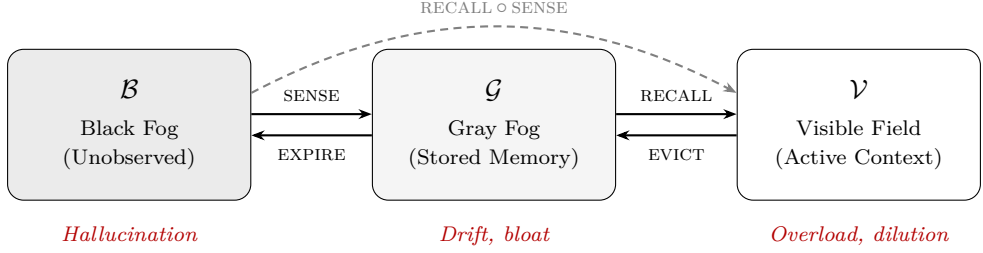

The composition $\textup{\textsc{recall}} \circ \textup{\textsc{sense}}$
maps directly from unobserved territory to the reasoning surface. In
minimal agent implementations, $\textup{\textsc{sense}}$ may deposit
results directly into $\mathcal{V}$ (bypassing $\mathcal{G}$
entirely). We treat direct $\mathcal{B} \to \mathcal{V}$ as a
collapsed composition of $\textup{\textsc{sense}}$ followed by
immediate $\textup{\textsc{recall}}$; Context Cartography asserts that
this composition should be mediated---raw sensing outputs should be
routed through $\mathcal{G}$ or at minimum transformed before entering
$\mathcal{V}$, to prevent visible field contamination. Mediation is
most critical when the sensing output is large relative to
$|\mathcal{V}|$ or when its format does not match the current
projection schema $p$; for small, pre-structured outputs (e.g., a
single API response returning typed fields), the collapsed
composition is acceptable. \Cref{fig:zones} illustrates the tripartite
zonal architecture and its transitions.

Each zone is characterized by a distinct failure mode when governance
is absent:

\paragraph{Black Fog ($\mathcal{B}$): Hallucination.}
When an agent generates assertions conditioned on $u \in \mathcal{B}$
as if $u \in \mathcal{V}$, the result is
hallucination~\citep{huang2023hallucination}. Although
\citet{kadavath2022know} show that models possess partial
self-knowledge about their uncertainty, this calibration is unreliable
at epistemic boundaries---models frequently generate confident
continuations consistent with pre-training statistics for information
they have not actually observed.

\paragraph{Gray Fog ($\mathcal{G}$): Drift and Bloat.}
Memory is inherently distinct from active visibility: the underlying
reality may have changed since last
observation~\citep{liu2025memorysurvey}. If $\mathcal{G}$ is
maintained as a chronological log, retrieval yields unstructured text
that overwhelms $\mathcal{V}$ upon
re-entry~\citep{packer2024memgpt}. \citet{ace2025} identify two specific failure modes: \emph{brevity
bias} (optimization toward short, generic outputs) and
\emph{context collapse} (monolithic rewriting eroding accumulated
knowledge). We recharacterize both as transition-level failures below.

\paragraph{Visible Field ($\mathcal{V}$): Overload and Dilution.}
The visible field is analogous to the episodic buffer in Baddeley's
working memory model~\citep{baddeley2000episodic}---a limited-capacity
interface between perception and long-term memory. Cognitive load
theory~\citep{sweller1988cognitive} predicts that exceeding this
capacity degrades performance. In LLMs, the corresponding failure
modes are \emph{attention dilution}~\citep{liu2024lost} and
\emph{constraint drift} (system instructions losing influence as
context grows~\citep{gu2026scaling}).

\paragraph{Transition-Level Failure Modes.}
The zone-level failure modes above arise from the \emph{state} of each
zone; a complementary class of failures arises from how information
\emph{moves between} zones:
\begin{itemize}[leftmargin=*, nosep]
  \item \textbf{Visible field contamination.} When the composition
        $\textup{\textsc{recall}} \circ \textup{\textsc{sense}}$ is
        executed without mediation---raw tool output or search results
        entering $\mathcal{V}$ without simplification ($\phi$) or
        format transformation ($\pi^{+}$)---the visible field absorbs
        noise at the expense of signal. This is not a missing
        operator but a \emph{failure to compose} existing operators on
        the inbound path.
  \item \textbf{Context collapse via destructive compaction.} If
        $\pi^{-}$ is applied without first archiving the original
        content in $\mathcal{G}$, the pre-compaction context
        transitions directly to $\mathcal{B}$ (permanently lost)
        rather than $\mathcal{G}$ (retrievable). The
        \emph{context collapse} and \emph{brevity bias} identified by
        \citet{ace2025} are precisely this failure: $\phi$ or $\pi^{-}$
        applied destructively, bypassing archival.
  \item \textbf{Cross-agent projection incoherence.} In multi-agent
        settings, each agent maintains its own zonal structure. When
        agents exchange information, they implicitly apply $\pi^{+}$
        across agent boundaries. If their projection schemas are
        incompatible---one agent's summary omits distinctions another
        agent requires---the receiving agent's $\mathcal{V}$ contains
        well-formed but semantically distorted content. This
        is supported by systematic failure analyses:
        \citet{cemri2025whyfail} identify \emph{deficient theory of
        mind} as a distinct inter-agent failure category in which
        agents fail to model each other's informational needs, and
        \citet{kostka2025synergy} show that integrating Theory of Mind
        mechanisms improves multi-agent reasoning.
  \item \textbf{Compound failure in iterative workflows.} When an
        agent revisits an evolving artifact across sessions (e.g.,
        multi-session debugging), failures compound: compacted memory
        does not distinguish original content from recent changes
        (missing $\lambda$), discarded diff metadata prevents
        detecting what changed (destructive $\pi^{-}$), and the agent
        relies on stale memory rather than re-reading
        ($\sigma$ over $\rho$). This pattern arises naturally in any
        long-running workflow involving iterative revision.
\end{itemize}
These transition-level failures manifest as operator \emph{omission}
or \emph{miscomposition} rather than intrinsic zone pathologies---each
is preventable using the existing operator set, and as the compound
pattern shows, a single missing operator often triggers cascading
failures in others.

\Cref{tab:zones} summarizes the tripartite architecture.

\begin{table}[t]
\centering
\caption{The tripartite zonal architecture of contextual space.
Zone-level failure modes are listed here; transition-level failures
(contamination, destructive compaction, cross-agent incoherence,
compound iterative failure) are characterized in the text.}
\label{tab:zones}
\small
\begin{tabularx}{\textwidth}{@{} >{\raggedright\arraybackslash}p{1.8cm} X X >{\raggedright\arraybackslash}p{3.2cm} @{}}
\toprule
\textbf{Zone} & \textbf{AI Counterpart} & \textbf{Cartographic Function} & \textbf{Failure Mode} \\
\midrule
Black Fog $\mathcal{B}$
  & Tool use, search, code execution~\citep{yao2023react, wang2023voyager}
  & Reconnaissance, surveying
  & Hallucination~\citep{huang2023hallucination} \\[4pt]
Gray Fog $\mathcal{G}$
  & Structured memory, vector DBs~\citep{jing2024vecdb}, persisted state
  & Archival mapping, state tracking
  & Drift, staleness, bloat~\citep{xiong2025memory} \\[4pt]
Visible Field $\mathcal{V}$
  & Active context window, prompt assembly~\citep{khattab2023dspy}
  & Tactical projection surface
  & Overload, attention dilution~\citep{liu2024lost} \\
\bottomrule
\end{tabularx}
\end{table}

\subsection{Salience Geometry of the Visible Field}
\label{sec:geometry}

The visible field $\mathcal{V}$ is not a uniform surface. We
characterize its geometry through a salience function that captures how
information at different positions contributes to downstream reasoning.

\begin{definition}[Salience Function]
\label{def:salience}
For a context window of length $n$, the \emph{salience function}
$s: \{1, \ldots, n\} \times \mathbb{N} \to [0,1]$ assigns to each
position $i$ a weight reflecting its effective contribution to the
model's next-token prediction. Under ideal (uniform) attention,
$s(i,n) = 1/n$ for all $i$. Under empirical transformer
attention~\citep{vaswani2017attention}, $s$ exhibits a characteristic
U-shaped profile.
\end{definition}

\paragraph{Salience Non-Uniformity.}
\citet{liu2024lost} documented that $s$ peaks at the beginning
(primacy bias) and end (recency bias) of the sequence, with a trough
in intermediate positions. \citet{wu2025emergence} provide a
graph-theoretic proof that this non-uniformity arises from the
interaction of causal masking and rotary positional encodings, and
\citet{li2025lost} demonstrate a complementary ``lost in the distance''
effect: $s$ degrades as the token distance between related elements
increases, even when both remain within $\mathcal{V}$.
The problem compounds at scale. \citet{vasylenko2025sparse} show that
normalized attention entropy grows with $n$ as attention mass disperses
across an expanding context, establishing a fundamental limitation:
softmax-based transformers face increasing difficulty maintaining
signal fidelity as context length grows.

\paragraph{Capacity Constraints.}
The discrepancy between stated and effective context utilization is
severe. \citet{paulsen2025mecw} find that most models show significant
accuracy degradation by 1,000 tokens, utilizing less than 1\% of stated
capacity; \citet{wang2026degradation} identify critical thresholds
beyond which performance collapses catastrophically---a phenomenon they
term ``shallow long-context adaptation.'' These capacity limits are
compounded by the rigid append-only paradigm of the KV
cache~\citep{li2025kvcache}: if an agent discovers during
$\textup{\textsc{sense}}$ that a fact at position $k$ is incorrect, it cannot
update that belief without reprocessing the entire sequence from $k$
onward. The discovery of ``attention
sinks''~\citep{xiao2024streamingllm}---initial tokens absorbing
disproportionate attention regardless of content---confirms that
positional gradients create structural anchors that further constrain
effective capacity.

\begin{thesis}[Necessity of External Governance]
\label{prop:necessity}
Under the salience geometry characterized above, the visible field
$\mathcal{V}$ cannot be treated as a passive container. Deliberate
transformations are required to maintain reasoning coherence as
$|\mathcal{V}|$ grows. We term these transformations
\emph{cartographic operators}.
\end{thesis}

\emph{Argument.} The claim follows from the conjunction of four
empirically established properties. (P1)~Non-uniform $s$ means that
information placed at intermediate positions contributes less to
downstream predictions than information at the extremes; a passive
container offers no mechanism to ensure high-priority content occupies
high-salience positions. (P2)~Entropy growth means that as $n$
increases, each additional token dilutes attention over all prior
tokens; without active compression, signal-to-noise ratio degrades
monotonically. (P3)~The append-only KV cache means that incorrect or
stale information at position $k$ cannot be revised without
reprocessing from $k$ onward; a passive container has no mechanism to
retract or update prior state. (P4)~Effective capacity far below
nominal means that the model's ability to utilize context degrades
well before the stated window is filled; without active filtering,
the agent operates on a fraction of the information it nominally has
access to. Any system relying on passive concatenation inherits all
four failure modes simultaneously; therefore, deliberate
transformations---repositioning, compression, eviction,
filtering---are structurally required.

\paragraph{Cartographic Laws of Context Representation.}
The necessity argument of Thesis~\ref{prop:necessity} can be distilled
into three principles
that parallel foundational laws of cartographic
representation~\citep{snyder1987projections, maceachren1995maps}:
\begin{enumerate}[leftmargin=*, nosep]
  \item \textbf{Bounded Surface Law.} Reasoning requires a bounded
        visible field; not all information can be simultaneously
        active. This parallels the cartographic principle that maps are
        finite representations of unbounded territory---the
        cartographer must decide what to include, not merely how to
        render it.
  \item \textbf{Scale-Distortion Trade-off.} Any compression of
        information into a bounded representation introduces
        distortion; the governance task is to choose \emph{which}
        distortions are acceptable. In map projection, no single
        projection preserves angles, areas, and distances
        simultaneously~\citep{snyder1987projections}; in context
        management, no single transformation preserves all semantic
        properties simultaneously.
  \item \textbf{Salience Geometry Constraint.} The bounded surface
        exhibits non-uniform effectiveness across positions;
        information placement affects reasoning quality. This parallels
        the cartographic principle that map readability depends on
        feature placement (visual hierarchy, figure-ground
        separation), not just feature presence.
\end{enumerate}
The seven cartographic operators defined below are the transformations
that enforce these laws within and across epistemic zones.

\subsection{Cartographic Operators}
\label{sec:operators}

Traditional cartography recognizes that a map is never a 1:1
reproduction of territory---it is the result of deliberate operators
applied to raw terrain~\citep{mcmaster1992generalization,
roth2011cartographic}. We define seven operators for contextual space,
derived from a systematic analysis of all non-trivial transformations
within and between epistemic zones. For each, we specify: (a)~a formal
definition as a transformation, (b)~the zones and transitions it
governs, and (c)~whether the correspondence with its classical cartographic counterpart
is a \emph{structural isomorphism} (preserving the same mathematical
structure) or a \emph{functional analogy} (serving the same purpose
through different mechanisms).

\paragraph{Design Principles.}
The operator set is organized by two orthogonal criteria:
\emph{transformation type} (what the operator does to information) and
\emph{zone scope} (where it applies). Most operators are
\emph{zone-general}---they define a transformation type that applies
wherever information of the appropriate kind exists. One operator
($\delta$, displacement) is zone-specific, arising from positional
salience gradients present in sequential architectures. Three operators ($\rho$,
$\sigma$, and $\pi$) are \emph{boundary operators}, governing how
information crosses zone transitions: $\rho$ determines \emph{where}
to explore, $\sigma$ determines \emph{which} items cross a boundary,
and $\pi$ determines \emph{how} items are represented upon crossing.
For each boundary operator, the formal type signature
(e.g., $\rho: \mathcal{B} \to \mathcal{P}(\mathcal{B})$) captures the
\emph{decision} the operator makes, while the zone scope in
\Cref{tab:operators} indicates the \emph{transition} it governs.

\begin{definition}[Reconnaissance]
\label{def:reconnaissance}
$\rho: \mathcal{B} \to \mathcal{P}(\mathcal{B})$ determines which
elements of unobserved territory to explore (what tools to invoke,
what queries to issue, what code to execute). Governs
$\textup{\textsc{sense}}$ transitions. The defining characteristic is
decision under uncertainty: the agent does not know what the
exploration will yield.
\end{definition}

\begin{definition}[Selection]
\label{def:selection}
$\sigma: Z \to \mathcal{P}(Z)$ for $Z \in \{\mathcal{V},
\mathcal{G}\}$ determines which elements cross a zone boundary.
Governs $\textup{\textsc{recall}}$, $\textup{\textsc{evict}}$, and
$\textup{\textsc{expire}}$ transitions via three modes:
$\sigma_{\textup{recall}}: \mathcal{G} \to \mathcal{P}(\mathcal{G})$
(select what to project into $\mathcal{V}$),
$\sigma_{\textup{evict}}: \mathcal{V} \to \mathcal{P}(\mathcal{V})$
(select what to archive from $\mathcal{V}$),
$\sigma_{\textup{expire}}: \mathcal{G} \to \mathcal{P}(\mathcal{G})$
(enforce lifecycle policies). Formally, $\sigma$ is a policy over the
source zone's inventory; its output parameterizes whichever transition
it governs.
\end{definition}

Reconnaissance and selection are both boundary operators, but they
differ in the epistemic status of their input. $\rho$ operates under
uncertainty: the agent does not know what $\mathcal{B}$ contains and
must plan exploration based on predictions about what might be useful.
$\sigma$ operates over known inventory: the agent has metadata about
$\mathcal{G}$ (or direct access to $\mathcal{V}$) and can score items
by relevance. This distinction---planner over unknown territory vs.\
filter over known content---corresponds to the
exploration--exploitation boundary in decision
theory~\citep{cuconasu2024power, shi2023large}, and is tested
directly in \Cref{sec:benchmark}.

\begin{definition}[Simplification]
\label{def:simplification}
$\phi: Z \to Z'$ where $|Z'| < |Z|$ for zone $Z \in \{\mathcal{V},
\mathcal{G}\}$, subject to the constraint that task-relevant semantic
content of $Z$ is preserved in $Z'$. Zone-general.
\end{definition}

\begin{definition}[Aggregation]
\label{def:aggregation}
$\alpha: \{z_1, \ldots, z_k\} \to z'$ fuses $k$ semantically similar
elements into a single representative, for elements in zone
$Z \in \{\mathcal{V}, \mathcal{G}\}$. Zone-general.
\end{definition}

Simplification reduces individual elements (e.g., compressing verbose
tool output, condensing stored memory
entries~\citep{complexity2025}). What constitutes ``task-relevant
semantic content'' depends on the downstream task, which is why
simplification quality remains an empirical criterion rather than a
formal guarantee.
Aggregation merges multiple elements
(e.g., deduplicating redundant observations, fusing repeated
signals). Agentic memory systems implement aggregation
explicitly~\citep{xu2025amem}, multi-agent episodic
reconstruction~\citep{wang2026emem} extends this across agent
boundaries, and trajectory reduction
methods~\citep{agentdiet2025} illustrate the complementary role of
selection (removing redundant steps rather than fusing them). Both
simplification and aggregation apply within
$\mathcal{V}$ and within $\mathcal{G}$.

\begin{definition}[Projection]
\label{def:projection}
$\pi: (Z_{\text{src}}, p) \to Z_{\text{dst}}$ transforms information
across the $\mathcal{G} \leftrightarrow \mathcal{V}$ boundary according
to a projection schema $p = (f, m, r, d)$ specifying format~$f$,
modality~$m$, resolution~$r \in \{r_{\text{coarse}}, \ldots,
r_{\text{fine}}\}$, and structural dimensionality~$d$. Bidirectional:
$\pi^{+}: (\mathcal{G}, p) \to \mathcal{V}$ (forward projection) and
$\pi^{-}: (\mathcal{V}, p) \to \mathcal{G}$ (inverse projection,
widely known as compaction).
\end{definition}

Forward projection subsumes operations that reduce structural
dimensionality---such as the context compression strategies
of~\citet{kang2025acon} and the ``structure-then-select'' approach
of~\citet{zhou2025edu}, which decomposes text into a discourse-unit
tree and then selects query-relevant sub-trees for
linearization---and operations that adjust representational
resolution~\citep{liu2025memorysurvey, chen2025dast}. Both are
parameters of the projection schema rather than independent
operations. Every projection necessarily distorts: the choice of $p$
determines which properties are preserved and which are sacrificed. A
well-chosen projection maintains relational structure (adjacency,
containment, causal links) even as the surface representation changes.
A particularly severe distortion arises from modality mismatch:
content stored in one modality but projected in another (e.g., a
diagram rendered as text) may lose spatial structure even when
propositional content is preserved.

\begin{definition}[Displacement]
\label{def:displacement}
$\delta: (v, i) \mapsto (v, j)$ repositions element $v$ from position
$i$ to position $j$ in the sequence where $s(j,n) > s(i,n)$.
Zone-specific ($\mathcal{V}$ only): compensates for the non-uniform
salience function of \Cref{sec:geometry}.
\end{definition}

Cartographic displacement repositions map features to resolve visual conflicts
caused by finite spatial resolution; context displacement repositions
information to compensate for non-uniform attention
salience~\citep{liu2024lost, wu2025emergence, hsieh2024found}. Both
sacrifice positional accuracy for functional effectiveness, but
compensate for different medium constraints (2D overlap vs.\ 1D
salience gradients). Common implementation patterns include
constraint pinning (re-projecting invariant constraints into the first
$k$ tokens at each turn), recency injection (appending a rolling
summary of high-priority state to the sequence end), and
salience-aware assembly (ordering segments to align critical content
with primacy and recency peaks). Displacement is irreducible to selection: $\sigma$ decides
\emph{which} items enter $\mathcal{V}$; $\delta$ decides \emph{where}
they sit once inside.

\begin{definition}[Layering]
\label{def:layering}
$\lambda: Z \to Z_1 \times \cdots \times Z_m$ partitions a zone
$Z \in \{\mathcal{V}, \mathcal{G}\}$ into $m$ typed namespaces.
Zone-general.
\end{definition}

Within $\mathcal{V}$, layering separates system constraints, task
state, retrieved memory, and fresh observations into distinct semantic
layers with explicit priority ordering---\citet{wallace2024hierarchy}
demonstrate that training LLMs to respect a typed instruction
hierarchy (system $>$ user $>$ third-party) dramatically improves
robustness to prompt injection. Within $\mathcal{G}$, layering
organizes stored memory into typed containers (as in MemOS's MemCube
abstraction~\citep{memos2025}). Layering can be applied
recursively---each namespace itself partitioned into
sub-namespaces---producing the spatial hierarchies fundamental to both
cartography and context engineering (e.g., repo $\to$ package $\to$ module
$\to$ class $\to$ method in a codebase, or domain $\to$ concept $\to$
fact in a knowledge base). Recursive $\lambda$ is the formal mechanism
underlying hierarchical $\mathcal{G}$ structures such as OpenViking's
filesystem tree~\citep{openviking2025} and MemOS's nested graph
schemas~\citep{memos2025b}.

\subsection{Operator Properties and Cartographic Correspondences}
\label{sec:properties}

Each operator corresponds to a classical cartographic counterpart. We now characterize these correspondences, the invariants they preserve, and the points at which the geographic and contextual settings diverge.

\paragraph{Cartographic Correspondences.}
Each operator has a counterpart in cartographic generalization,
classified as either a structural isomorphism or a functional analogy.
The classification criterion is: a correspondence is an
\emph{isomorphism} when the cartographic and context operators share
input/output type structure and composition behavior (so that
theorems about one transfer to the other), and an \emph{analogy} when
they serve the same functional role through structurally different
mechanisms (so that design intuitions transfer but formal results do
not).
$\sigma$, $\alpha$, $\pi$, and $\lambda$ are structural
isomorphisms: $\sigma$ preserves relevance ranking, $\alpha$ preserves
semantic equivalence classes, $\pi$ preserves relational structure
across representation change, and $\lambda$ preserves namespace
disjointness. $\phi$ and $\delta$ are functional analogies:
cartographic simplification reduces geometric complexity while context
simplification reduces token count~\citep{complexity2025};
cartographic displacement resolves visual overlap while context
displacement
compensates for salience gradients.
\Cref{tab:operators} records the correspondence type for each
operator; \Cref{tab:gis_mapping} provides the detailed mapping from
classical cartographic operators to their context counterparts.

\paragraph{Representational Invariants.}
In classical cartography, map projections are classified by the
property they preserve: conformality (local shape), equivalence
(area), or equidistance (distance)~\citep{snyder1987projections}. A
foundational result is that these are mutually exclusive---no
projection preserves all three simultaneously. We identify an
analogous triad for context transformations:
\begin{itemize}[leftmargin=*, nosep]
  \item \textbf{Semantic invariance}: the meaning of individual
        elements is preserved despite surface-level reduction.
        Primarily enforced by $\phi$ (simplification).
  \item \textbf{Structural invariance}: relational structure
        (adjacency, containment, causal links) is preserved despite
        representational change. Primarily enforced by $\pi$
        (projection).
  \item \textbf{Salience invariance}: reasoning relevance is
        preserved despite positional change. Primarily enforced by
        $\delta$ (displacement).
\end{itemize}
As with map projections, no single context transformation preserves
all three invariants simultaneously: simplification sacrifices
structural detail, projection sacrifices surface semantics, and
displacement sacrifices chronological ordering. The existence of
multiple operators reflects this fundamental trade-off.

\paragraph{Distinguishing $\phi$, $\alpha$, and $\pi$.}
An apparent redundancy deserves explicit justification: $\phi$
(simplification, \Cref{def:simplification}), $\alpha$ (aggregation,
\Cref{def:aggregation}), and $\pi$ (projection,
\Cref{def:projection}) all transform representations, and one might argue they should be unified
into a single ``transformation'' operator. We maintain the three-way
split because each preserves a \emph{distinct invariant}, fails in a
\emph{distinct mode}, and corresponds to a \emph{distinct classical cartographic
operator}:
\begin{itemize}[leftmargin=*, nosep]
  \item $\phi$ preserves \textbf{semantic content} while reducing
        surface tokens. Input: one element. Output: one shorter
        element. Failure mode: critical details
        dropped~\citep{ace2025, complexity2025}. Cartographic analogue:
        simplifying a coastline's vertex count while preserving its
        shape~\citep{roth2011cartographic}.
  \item $\alpha$ preserves the \textbf{equivalence class} while
        reducing element count. Input: $k$ similar elements. Output:
        one composite representative. Failure mode: distinguishing
        features between merged elements lost. Cartographic analogue: merging
        individual buildings into a city-block
        polygon~\citep{mcmaster1992generalization}.
  \item $\pi$ preserves \textbf{relational structure} while changing
        the representational system. Input: one element in format $A$.
        Output: one element in format $B$. Failure mode: structural
        information lost in format
        translation~\citep{zhou2025edu}. Cartographic analogue:
        projecting coordinates from sphere to plane.
\end{itemize}
Collapsing these into a single ``transformation'' operator would
prevent diagnosing which specific invariant was violated when a
transformation fails.

\begin{table}[t]
\centering
\caption{Mapping from classical cartographic generalization operators
to context cartography operators. ``Isomorphism'' indicates shared
formal structure; ``Analogy'' indicates shared purpose with different
mechanism; ``Subsumed'' indicates the cartographic operator maps to a mode or
parameter of a context operator rather than a standalone counterpart.}
\label{tab:gis_mapping}
\small
\begin{tabularx}{\textwidth}{@{} >{\raggedright\arraybackslash}p{2.0cm}
  >{\raggedright\arraybackslash}p{2.7cm}
  l l
  X @{}}
\toprule
\textbf{Cartographic Op.} & \textbf{Classical Function} & \textbf{Context Op.}
  & \textbf{Corr.} & \textbf{Notes} \\
\midrule
Selection / elimination
  & Filter features by relevance
  & $\sigma$
  & Iso.
  & Both filter known inventory by task criteria \\[3pt]
Simplification
  & Reduce geometric complexity
  & $\phi$
  & Ana.
  & Cartographic: vertex count; Context: token count \\[3pt]
Aggregation
  & Merge similar features
  & $\alpha$
  & Iso.
  & Both fuse proximate items into composites \\[3pt]
Typification
  & Replace dense set with representatives
  & $\sigma$ (mode)
  & Sub.
  & Pattern-preserving sampling; a selection
    strategy \\[3pt]
Displacement
  & Resolve visual conflicts
  & $\delta$
  & Ana.
  & Cartographic: 2D overlap; Context: 1D salience \\[3pt]
Collapse
  & Reduce dimensionality
  & $\pi$ (param.)
  & Sub.
  & Dimensionality reduction within projection
    schema \\[3pt]
Symbolization
  & Assign typed visual categories
  & $\lambda$
  & Iso.
  & Both enforce semantic namespace separation \\[3pt]
Map projection
  & Transform between coordinate systems
  & $\pi$
  & Iso.
  & Both transform representations with controlled
    distortion \\[3pt]
\midrule
\multicolumn{5}{@{}l}{\emph{Context operators without classical
  cartographic generalization counterpart:}} \\[3pt]
Reconnaissance
  & (Field surveying / data acquisition)
  & $\rho$
  & ---
  & Precedes generalization; cartographic analogue
    is survey planning \\
\bottomrule
\end{tabularx}
\end{table}

\paragraph{Where Correspondences Diverge.}
The isomorphisms and analogies in \Cref{tab:gis_mapping} hold at the
level of formal structure, but the underlying media differ in ways
that affect operator behavior. Three systematic divergences are worth
noting. First, cartography operates on \emph{continuous 2D space} with
visual perception constraints, while context operates on \emph{discrete
1D sequences} with attention constraints; this means cartographic
displacement
resolves pairwise feature conflicts (a local, geometric problem),
whereas context displacement compensates for a global salience
gradient imposed by the architecture. Second, cartographic
generalization is typically performed \emph{offline} during map
compilation, while context operators must execute \emph{online}
within the latency budget of each agent turn---imposing computational
constraints that classical cartographic operators do not face. Third,
geographic features have stable
identities across scales (a river is the same river at 1:50{,}000 and
1:1{,}000{,}000), whereas context elements may lose identity through
compaction: a summarized conversation turn is not the ``same'' element
at a different resolution but a new synthetic artifact. These
divergences do not invalidate the correspondences but constrain how
directly cartographic algorithms can be transferred to context
engineering.

\subsection{Composition and Architecture}
\label{sec:composition}

We now show that the seven operators cover all zonal transformations, compose into pipelines, and integrate into a minimal reference architecture.

\paragraph{Coverage.}
The seven operators cover all non-trivial transformations across the
zonal architecture:
\begin{itemize}[leftmargin=*, nosep]
  \item $\mathcal{B} \to \mathcal{G}$: governed by $\rho$
        (reconnaissance)
  \item $\mathcal{G} \to \mathcal{V}$: governed by
        $\sigma$ (selection) $+$ $\pi^{+}$ (forward projection)
  \item $\mathcal{V} \to \mathcal{V}$: governed by $\phi$, $\alpha$,
        $\delta$, $\lambda$
  \item $\mathcal{V} \to \mathcal{G}$: governed by
        $\sigma$ (select what to evict) $+$ $\pi^{-}$ (compaction)
  \item $\mathcal{G} \to \mathcal{G}$: governed by $\phi$, $\alpha$,
        $\lambda$ applied within $\mathcal{G}$ (memory consolidation)
  \item $\mathcal{G} \to \mathcal{B}$: governed by $\sigma$
        (lifecycle-aware expiration policy)
\end{itemize}
\noindent This coverage claim, along with the zonal partition and
layering disjointness properties, has been mechanically verified in
Lean~4 (\Cref{app:lean}).

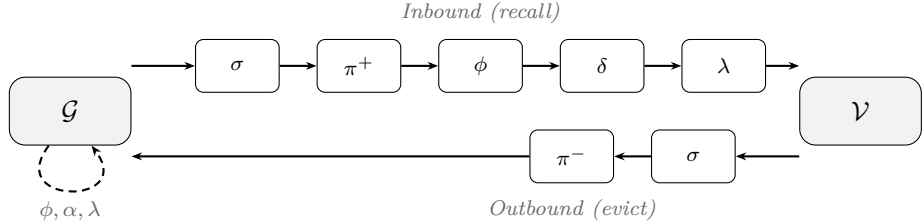
\begin{figure}[t]
\centering
\begin{tikzpicture}[
  op/.style={draw, rounded corners=3pt, minimum height=0.7cm,
    minimum width=1.1cm, align=center, font=\scriptsize,
    fill=white},
  store/.style={draw, rounded corners=6pt, minimum height=0.9cm,
    minimum width=1.6cm, align=center, font=\small, fill=black!5},
  arr/.style={-{Stealth[length=4pt]}, thick},
  lab/.style={font=\scriptsize\itshape, text=black!60},
  >=Stealth
]
\node[store] (G) at (0, 0) {$\mathcal{G}$};
\node[store] (V) at (10.4, 0) {$\mathcal{V}$};

\node[op] (s1) at (2.2, 0.6) {$\sigma$};
\node[op] (p1) at (3.8, 0.6) {$\pi^{+}$};
\node[op] (ph) at (5.4, 0.6) {$\phi$};
\node[op] (de) at (7.0, 0.6) {$\delta$};
\node[op] (la) at (8.6, 0.6) {$\lambda$};

\draw[arr] (G.east |- s1) -- (s1);
\draw[arr] (s1) -- (p1);
\draw[arr] (p1) -- (ph);
\draw[arr] (ph) -- (de);
\draw[arr] (de) -- (la);
\draw[arr] (la) -- (V.west |- la);

\node[lab, above=2pt of ph] {Inbound (recall)};

\node[op] (s2) at (8.2, -0.6) {$\sigma$};
\node[op] (p2) at (6.6, -0.6) {$\pi^{-}$};

\draw[arr] (V.west |- s2) -- (s2);
\draw[arr] (s2) -- (p2);
\draw[arr] (p2) -- (G.east |- p2);

\node[lab, below=2pt of p2] {Outbound (evict)};

\draw[arr, densely dashed] ([xshift=-8pt]G.south)
  .. controls +(-0.6,-0.8) and +(0.6,-0.8) ..
  node[below, lab] {$\phi, \alpha, \lambda$}
  ([xshift=8pt]G.south);

\end{tikzpicture}
\caption{Cartographic pipelines.  The inbound pipeline
$\lambda \circ \delta \circ \phi \circ \pi^{+} \circ \sigma$
transforms content from $\mathcal{G}$ to $\mathcal{V}$; the outbound
pipeline $\pi^{-} \circ \sigma$ archives from $\mathcal{V}$ to
$\mathcal{G}$; the dashed loop represents asynchronous maintenance
within $\mathcal{G}$.}
\label{fig:pipeline}
\end{figure}

\paragraph{Completeness.}
Several prima facie gaps in the operator set reduce, on analysis, to
failure modes of existing operators or their compositions rather than
to missing primitives:
\begin{itemize}[leftmargin=*, nosep]
  \item \textbf{Output gating} (mediating raw $\textup{\textsc{sense}}$
        output before it enters $\mathcal{V}$) is not a separate
        operator but the prescribed application of $\phi \circ \pi^{+}$
        on the inbound path; its absence produces visible field
        contamination.
  \item \textbf{Multi-agent context packaging} (an orchestrator
        composing a context payload for a subagent) is a composition
        $\lambda \circ \pi^{+} \circ \sigma$ applied to the
        orchestrator's own $\mathcal{V}$ or $\mathcal{G}$, with the
        subagent's initial context as destination. Poor packaging is a
        failure of $\sigma$ (wrong selection), $\pi^{+}$ (wrong
        format), or $\lambda$ (missing namespace separation).
  \item \textbf{Faithfulness verification} (checking that a
        transformation preserved task-relevant semantics) is a quality
        criterion for operators, not a transformation itself---analogous
        to map validation in cartography, which evaluates generalization
        operators but is not itself a generalization operator.
  \item \textbf{In-place context reset} (replacing an entire
        $\mathcal{V}$ with a condensed summary, as in conversation
        compaction) decomposes into a \emph{compaction cycle}:
        $\pi^{-}$ (archive original to $\mathcal{G}$) followed by
        $\sigma \circ \pi^{+}$ (select and project summary back into
        $\mathcal{V}$). We reserve \emph{compaction} for the $\pi^{-}$
        step alone (not the full cycle). When the archival step is
        skipped, the original is
        lost to $\mathcal{B}$---producing context collapse.
\end{itemize}
\paragraph{Composition.}
Operators compose into \emph{cartographic pipelines}. The inbound
pipeline from gray fog to visible field applies $\sigma$ (select what
to recall), then $\pi^{+}$ (project into token format), then $\phi$
(simplify), then $\delta$ (position for salience), then $\lambda$
(assign to namespace):
$\lambda \circ \delta \circ \phi \circ \pi^{+} \circ \sigma$.
The outbound pipeline reverses this: $\sigma$ (select what to evict),
then $\pi^{-}$ (compact into storage format):
$\pi^{-} \circ \sigma$.
A maintenance cycle applies $\{\phi, \alpha, \lambda\}$ within
$\mathcal{G}$ asynchronously---the formal characterization of
processes such as Letta's sleep-time memory
reorganization~\citep{letta2025context}.
Two basic composition properties are worth noting.
First, $\phi$ is approximately idempotent: applying simplification
twice yields marginal further reduction, since most redundancy is
removed in the first pass. Second, $\delta$ and $\phi$ do not
commute in general: simplifying before repositioning ($\delta \circ
\phi$) operates on already-compressed content, while repositioning
before simplifying ($\phi \circ \delta$) may discard content that was
just moved to a high-salience position. The prescribed pipeline order
places $\phi$ before $\delta$ to avoid this interaction. A
systematic analysis of commutativity and idempotency across all
operator pairs remains an open problem (\Cref{sec:open}).
\Cref{fig:pipeline} illustrates the inbound, outbound, and
maintenance pipelines.

\paragraph{Scale Modulation.}
In cartography, \emph{map scale} is the ratio between a distance on the map
and the corresponding distance on the ground; it determines which
features are representable and which operators apply. We define
\emph{contextual scale} analogously as the ratio between the
information content of a source entity and its representation in
$\mathcal{V}$. At coarse contextual scale, a 500-line module is
represented by a one-sentence summary; at fine contextual scale, the
full source is projected. Contextual scale is not a single operator
but a \emph{coordination policy} that adjusts all operator parameters
simultaneously~\citep{mcmaster1992generalization, roth2011cartographic}:
reducing the scale increases simplification intensity, triggers
aggregation, broadens selection scope, and may suppress entire
layers. Prior work on multi-resolution context
management---including hierarchical
compression~\citep{kang2025acon}, discourse-unit
decomposition~\citep{zhou2025edu}, and tiered memory
loading~\citep{openviking2025}---addresses individual resolution
choices within a single operator. Each ``zoom level,'' however, does
not merely change $\pi$'s resolution parameter $r$---it changes the
parameters of $\sigma$, $\phi$, $\alpha$, and $\lambda$ in concert.
Scale modulation is therefore not an eighth operator but a policy
over the pipeline's parameterization, governing how operator settings
co-vary with a target resolution. In practice, the resolution levels
available to this policy correspond to levels in a spatial hierarchy:
zooming in descends the hierarchy, zooming out ascends it.
Aggregation ($\alpha$) constructs these hierarchies---merging sibling
elements into their parent level---while scale modulation navigates
them. Among the systems in \Cref{sec:practices}, OpenViking and Letta
implement explicit scale policies (tiered loading and core/recall
separation, respectively), while Claude Code makes per-operation
resolution choices without a coordinating policy.

\paragraph{Minimal Reference Architecture.}
A cartographically governed agent maintains a structured
$\mathcal{G}$ and a context assembler that builds $\mathcal{V}$ at
each turn by applying the inbound pipeline, the outbound pipeline,
and asynchronous $\mathcal{G}$-maintenance as defined above. A
decision layer at the $\rho$/$\sigma$ boundary routes knowledge gaps
to tool invocation or memory recall based on the agent's epistemic
state.

\Cref{tab:operators} summarizes the seven operators.

\begin{table}[t]
\centering
\caption{Cartographic operators for contextual space transformation.}
\label{tab:operators}
\small
\begin{tabularx}{\textwidth}{@{}l X l l@{}}
\toprule
\textbf{Operator} & \textbf{Formal Role} & \textbf{Zone Scope} & \textbf{Cart.\ Corr.} \\
\midrule
Reconnaissance $\rho$   & Determine what to explore
                        & $\mathcal{B} \to \mathcal{G}$
                        & --- \\
Selection $\sigma$      & Filter known inventory by relevance
                        & $\mathcal{G} \leftrightarrow \mathcal{V}$
                        & Isomorphism \\
Simplification $\phi$   & Reduce tokens, preserve structure
                        & $\mathcal{V}$ or $\mathcal{G}$
                        & Analogy \\
Aggregation $\alpha$    & Fuse repeated signals
                        & $\mathcal{V}$ or $\mathcal{G}$
                        & Isomorphism \\
Projection $\pi$        & Transform representation across boundary
                        & $\mathcal{G} \leftrightarrow \mathcal{V}$
                        & Isomorphism \\
Displacement $\delta$   & Reposition for salience
                        & $\mathcal{V}$ only
                        & Analogy \\
Layering $\lambda$      & Separate into typed namespaces
                        & $\mathcal{V}$ or $\mathcal{G}$
                        & Isomorphism \\
\bottomrule
\end{tabularx}
\end{table}

\paragraph{Mapping to Engineering Vocabulary.}
\Cref{tab:mapping} connects the framework to common agent engineering
concepts, showing which zones, transitions, and operators each concept
involves.

\begin{table}[t]
\centering
\caption{Common agent engineering concepts mapped to the cartographic
framework.}
\label{tab:mapping}
\small
\begin{tabularx}{\textwidth}{@{} >{\raggedright\arraybackslash}p{2.4cm}
  >{\raggedright\arraybackslash}p{3.0cm}
  X @{}}
\toprule
\textbf{Concept} & \textbf{Zone / Transition} & \textbf{Operators} \\
\midrule
Tool / function call
  & $\textup{\textsc{sense}}$: $\mathcal{B} \to \mathcal{G}$
  & $\rho$ (decide to call); $\phi \circ \pi^{+}$ (mediate result
    before entering $\mathcal{V}$) \\[3pt]
RAG retrieval
  & $\textup{\textsc{recall}}$: $\mathcal{G} \to \mathcal{V}$
  & $\sigma$ (query indexed corpus); $\pi^{+}$ (format chunks for
    $\mathcal{V}$) \\[3pt]
MCP server~\citep{hou2025mcp}
  & Standardized $\textup{\textsc{sense}}$ interface
  & $\rho$ (select server/tool); $\pi$ (common projection schema
    across providers) \\[3pt]
System prompt~\citep{mu2025systemprompt}
  & Pinned region of $\mathcal{V}$
  & $\lambda$ (typed namespace); $\delta$ (constraint pinning at
    primacy position) \\[3pt]
Conversation history~\citep{wang2023recursive}
  & $\mathcal{V}$ (live); $\mathcal{G}$ (after compaction)
  & $\pi^{-}$ (compaction cycle); $\sigma_{\textup{recall}}$
    (re-project from archive) \\[3pt]
Memory system~\citep{sumers2023coala, pink2025episodic}
  & $\mathcal{G}$
  & $\lambda$ (episodic / semantic / procedural layers);
    $\phi$, $\alpha$ (maintenance) \\[3pt]
Subagent delegation~\citep{zhang2025agentorchestra}
  & $\mathcal{B}$ from orchestrator's view
  & $\rho$ (delegate exploration); $\lambda$ (package context);
    $\phi \circ \pi^{+}$ (receive condensed result) \\[3pt]
Summarization
  & $\mathcal{V} \to \mathcal{V}$ or $\mathcal{V} \to \mathcal{G}$
  & $\phi$ (within-zone); $\pi^{-}$ (cross-zone compaction) \\[3pt]
Context window management
  & Cycle: $\mathcal{V} \to \mathcal{G} \to \mathcal{V}$
  & Full compaction cycle: $\sigma \circ \pi^{-}$ (archive),
    then $\sigma \circ \pi^{+}$ (re-project summary) \\
\bottomrule
\end{tabularx}
\end{table}

\section{Case Studies: Emergent Cartographic Practices}
\label{sec:practices}

We analyze four contemporary systems to demonstrate the
\emph{descriptive utility} of the framework---its ability to capture,
in a unified vocabulary, design patterns that these systems developed
independently for different use cases. To ensure reproducibility, we
evaluate each system--operator pair against a five-criterion rubric
measuring implementation depth (\Cref{sec:convergence}).

An important caveat: the operators in \Cref{sec:operators} were
informed in part by observing these systems, so the case studies are
not an independent test of the framework's predictive power. They
demonstrate that the vocabulary is descriptively adequate---that it
can account for the design choices these systems make---but not that
it predicted those choices. Predictive validation requires testing
against systems not used during framework development, which we
identify as a priority for future work.

\subsection{Claude Code: Subagent Isolation}
\label{sec:claude}

Claude Code~\citep{anthropic2025claudecode} implements cartographic
\emph{zoning}: it spawns subagents with bounded exploration contexts,
separating $\textup{\textsc{sense}}$ operations from the primary
$\mathcal{V}$. The orchestrator delegates $\rho$ (reconnaissance) to
subagents, which traverse $\mathcal{B}$ in isolation, then apply
$\phi$ (simplification) and $\pi^{-}$ (compaction) before the
orchestrator projects condensed results back into its visible field
via $\pi^{+}$. This achieves a weaker form of $\delta$ (displacement) through
salience-advantaged \emph{placement} rather than active repositioning:
summaries enter at the recency-biased end of the sequence, benefiting
from positional salience without explicit governance. The primary gap
is explicit resolution control within $\pi$---subagent summaries are
returned at a single projection schema.

The key architectural contribution is the \emph{agentic handoff}: the
orchestrator must compose a context package for each subagent via
$\lambda$ (layering), transferring task-relevant constraints without
projecting the entire global state.

\subsection{Letta (MemGPT): Memory as Operating System}
\label{sec:letta}

Letta~\citep{packer2024memgpt} restructures $\mathcal{G}$ as an
operating-system memory hierarchy. \emph{Core memory blocks} are
perpetually projected into $\mathcal{V}$ via fixed $\pi^{+}$
(implementing persistent $\textup{\textsc{recall}}$), while \emph{recall
memory} manages overflow through recursive $\pi^{-}$
(compaction)---evicted content is condensed and archived. Background
``sleep-time'' processes implement $\mathcal{G}$-internal maintenance
by applying $\{\phi, \alpha, \lambda\}$ within $\mathcal{G}$,
periodically reorganizing stored memory into hierarchical structures.
Context repositories~\citep{letta2025context} extend $\mathcal{G}$
with git-based versioning for persistent state. The core/recall tier
separation implements resolution control within $\pi$: core memory is
projected at full resolution, recall memory at reduced resolution.

\subsection{MemOS: Graph-Structured Memory}
\label{sec:memos}

MemOS~\citep{memos2025, memos2025b} treats $\mathcal{G}$ as
enterprise-grade infrastructure. The MemCube abstraction implements
$\lambda$ (layering) within $\mathcal{G}$ by isolating distinct
knowledge bases as composable, typed containers. DAG-based scheduling
implements $\sigma$ (selection) across multi-stage workflows. The
``next-scene prediction'' mechanism~\citep{memos2025b} implements
predictive $\sigma$ and $\pi^{+}$---preloading memory at appropriate
resolution before explicit request. The graph-structured storage
enables rich $\pi^{+}$ projection schemas: the same underlying memory
can be projected as entity lists, relationship summaries, or full
subgraph traversals depending on the task. Empirical results show a
159\% gain in temporal reasoning and 60.95\% token reduction on the
LoCoMo benchmark (self-reported).

\subsection{OpenViking: Hierarchical Context}
\label{sec:openviking}

OpenViking~\citep{openviking2025} spatializes $\mathcal{G}$ as a
hierarchical filesystem with a custom URI scheme. Its tiered loading
protocol (L0: ${\sim}$100 tokens, L1: ${\sim}$1,000 tokens, L2: full
content) is a direct implementation of multi-resolution $\pi^{+}$
(forward projection with explicit resolution parameter). Directory
recursion implements structured $\sigma$ (selection) via path-based
traversal rather than flat similarity search. The system preserves and
visualizes the ``retrieval trajectory,'' implementing a structural
reduction within $\pi^{+}$ by presenting the navigated path rather
than the full filesystem.

\subsection{Cross-System Analysis}
\label{sec:convergence}

To move beyond binary present/absent judgments, we evaluate each
system--operator pair on five criteria, each scored 0 or 1:
\begin{enumerate}[leftmargin=*, nosep]
  \item \textbf{Present (P)}: any mechanism performs this transformation.
  \item \textbf{Explicit (E)}: the mechanism is a named architectural
        component, not an emergent side-effect.
  \item \textbf{Configurable (C)}: the operator's parameters can be
        tuned by the user or developer.
  \item \textbf{Automated (A)}: the transformation occurs without
        manual intervention during normal operation.
  \item \textbf{Documented (D)}: the mechanism is described as a
        feature in official documentation or papers.
\end{enumerate}
Each cell in \Cref{tab:convergence} reports a score from 0 (absent)
to 5 (fully realized). Assignments are \emph{evidence-based lower
bounds}: they reflect what can be determined from public documentation,
technical papers, and source code (where available), and may
understate internal capabilities not externally visible. A key
scoring distinction: emergent side-effects (e.g., content entering at
the recency-biased end of an append-only sequence) may satisfy
\textbf{P} (Present) but not \textbf{E} (Explicit), since no
deliberate architectural mechanism performs the transformation. This
is why $\delta$ scores remain low despite every append-only system
exhibiting incidental positional bias.

\begin{table}[t]
\centering
\caption{Operator implementation depth across four systems. Each cell
scores five binary criteria (Present, Explicit, Configurable,
Automated, Documented; max = 5). Row means indicate per-operator
adoption; column means indicate per-system coverage.}
\label{tab:convergence}
\small
\begin{tabularx}{\textwidth}{@{} l *{4}{>{\centering\arraybackslash}X}
  >{\centering\arraybackslash}p{1.2cm} @{}}
\toprule
\textbf{Operator}
  & \textbf{Claude Code}
  & \textbf{Letta}
  & \textbf{MemOS}
  & \textbf{OpenViking}
  & \textbf{Mean} \\
\midrule
Reconnaissance $\rho$
  & 5 & 1 & 1 & 1 & 2.00 \\
Selection $\sigma$
  & 2 & 2 & 5 & 5 & 3.50 \\
Simplification $\phi$
  & 4 & 4 & 2 & 1 & 2.75 \\
Aggregation $\alpha$
  & 1 & 2 & 4 & 1 & 2.00 \\
Projection $\pi$
  & 2 & 5 & 5 & 5 & 4.25 \\
Displacement $\delta$
  & 2 & 1 & 1 & 1 & 1.25 \\
Layering $\lambda$
  & 5 & 5 & 5 & 5 & 5.00 \\
\midrule
\textbf{System mean}
  & 3.00 & 2.86 & 3.29 & 2.71 & 2.96 \\
\bottomrule
\end{tabularx}
\end{table}

Four quantitative observations emerge from the scored analysis.
First, $\lambda$ (layering) achieves a perfect mean score of 5.00---it
is the only operator fully realized as a primary, configurable,
automated, and documented mechanism in every system. This confirms
that namespace contamination is the most universally recognized failure
mode~\citep{ace2025}. Second, $\pi$ (projection) scores 4.25,
the highest among non-universal operators, indicating strong
convergence on the need for representational transformation at the
$\mathcal{G} \leftrightarrow \mathcal{V}$ boundary. Third, $\delta$
(displacement) scores only 1.25---the lowest of any operator---despite
addressing the best-documented failure
mode~\citep{liu2024lost}. Where it exists, it is implicit (e.g.,
summaries entering at the recency-biased end of the sequence in Claude
Code) rather than an explicit architectural mechanism. Fourth, the
systems bifurcate along the $\rho$/$\pi$ axis: Claude Code
emphasizes exploration governance ($\rho = 5$) while MemOS and
OpenViking emphasize projection governance ($\pi = 5$, $\sigma = 5$).
Letta bridges both strategies with strong projection ($\pi = 5$) and
moderate simplification ($\phi = 4$). The overall system mean of 2.96
out of 5.00 indicates that current systems implement approximately
60\% of the cartographic operator space, with systematic gaps in
reconnaissance ($\rho$), aggregation ($\alpha$), and displacement
($\delta$).

The transition-level failure modes of \Cref{sec:zones} map onto these
gaps: low $\phi$ scores expose systems to visible field contamination,
low $\rho$ scores leave the exploration--exploitation boundary
ungoverned, and the universal adoption of $\lambda$ (score~=~5 across
all systems) reinforces the namespace contamination finding above.

\section{Research Agenda}
\label{sec:agenda}

The preceding framework makes testable claims: that cartographic
operators improve reasoning coherence, and that their effects are
attributable to compensating for specific geometric properties of
the visible field. This section derives falsifiable predictions from
the framework and proposes a diagnostic benchmark for community
evaluation.

\subsection{Proposed Benchmark: Context Cartography Diagnostic (CCD)}
\label{sec:benchmark}

We propose a diagnostic benchmark with task categories designed to
isolate individual operators and zone transitions. Unlike
general-purpose long-context benchmarks~\citep{paulsen2025mecw}, the
CCD is designed to measure \emph{operator-specific} effects through
targeted task design:

\begin{itemize}[leftmargin=*]
  \item \textbf{Reconnaissance vs.\ selection tasks.} The agent must
        decide when to invoke tools ($\rho$: explore $\mathcal{B}$)
        versus answering from existing memory ($\sigma$: recall from
        $\mathcal{G}$), measuring the exploration--exploitation
        boundary. The agent is penalized both for unnecessary tool
        calls (wasteful $\rho$) and for hallucinating answers that
        could have been grounded by $\rho$.
  \item \textbf{Projection tasks.} The agent receives information in
        various storage-native formats (JSON, graph structures,
        hierarchical file listings) and must project it into effective
        reasoning context. Tasks require switching between
        summary-level and detail-level reasoning within a single
        session (e.g., first identify the relevant module at coarse
        resolution, then debug a specific function at fine resolution),
        testing multi-resolution $\pi$.
  \item \textbf{Displacement tasks.} Critical constraints (safety
        rules, formatting requirements) are placed at varying
        positions within the context. The agent must adhere to these
        constraints regardless of position. This extends the
        GM-Extract protocol~\citep{gupte2025gm} with agentic
        evaluation.
  \item \textbf{Simplification tasks.} The agent receives either raw
        tool output (verbose logs, full code listings) or simplified
        summaries, and must complete a downstream task. This measures
        whether $\phi$ preserves task-relevant information.
  \item \textbf{Aggregation tasks.} The agent receives $k$ partially
        overlapping observations (e.g., multiple search results
        covering the same topic, or repeated tool outputs with minor
        variations) and must produce a unified answer. Without
        $\alpha$, redundancy accumulates in $\mathcal{V}$, consuming
        token budget and diluting attention across repeated content.
  \item \textbf{Layering tasks.} Conflicting information is placed
        in different semantic layers (system prompt vs.\ retrieved
        memory vs.\ user input). The agent must correctly prioritize
        based on layer type.
\end{itemize}

The benchmark is designed for evaluation via operator ablation:
implementing the cartographic pipeline as composable middleware
layers and systematically removing individual operators to measure
their isolated contribution. Each configuration would be evaluated on
task accuracy, token consumption, and failure mode distribution
(hallucination rate, constraint adherence, information loss). Failure
mode labeling can be instrumented through tool-call logging (detecting
unnecessary $\rho$), gold-provenance comparison (distinguishing
hallucination from stale-memory reliance), and constraint-checker
oracles (measuring $\delta$/$\lambda$ adherence).

\subsection{Testable Predictions}
\label{sec:predictions}

The framework, combined with the transition-level failure analysis
(\Cref{sec:zones}), yields specific falsifiable predictions. Each
prediction identifies an operator, the failure mode expected when that
operator is absent, and the observable effect:

\begin{enumerate}[leftmargin=*]
  \item \textbf{Simplification necessity} ($\phi$). Removing
        simplification from the inbound pipeline increases visible
        field contamination: raw tool output displaces task-relevant
        content, degrading downstream accuracy. The effect should
        scale with tool output verbosity---high-verbosity tasks
        (full code listings, build logs) should show larger
        degradation than low-verbosity tasks (structured API
        responses).
  \item \textbf{Displacement--length interaction} ($\delta$). Removing
        salience-aware positioning increases constraint violation
        rates, with the effect size scaling with context length as the
        salience trough deepens~\citep{liu2024lost}. At short context
        lengths (below ${\sim}$4K tokens), displacement should have
        negligible effect; at long context lengths ($>$32K tokens),
        the effect should be substantial.
  \item \textbf{Layering--conflict interaction} ($\lambda$). Removing
        namespace separation increases layer-priority errors
        specifically when information sources conflict. On tasks with
        no inter-source conflicts, layering should have minimal
        effect; on tasks with deliberate conflicts (e.g., system
        prompt contradicts retrieved memory), the effect should be
        large.
  \item \textbf{Compaction mode} ($\pi^{-}$). Under archival
        compaction ($\pi^{-}$ with original preserved in
        $\mathcal{G}$), information loss should remain bounded over
        successive compaction cycles. Under destructive compaction
        ($\pi^{-}$ without archival), information loss should
        accumulate, with the gap widening over iterations---the
        quantitative signature of context
        collapse~\citep{ace2025}.
  \item \textbf{Exploration--exploitation boundary} ($\rho$/$\sigma$).
        Agents without explicit reconnaissance governance should
        exhibit a bimodal failure distribution: over-reliance on
        $\mathcal{G}$ (hallucination from stale memory when
        $\rho$ is under-used) or over-exploration of $\mathcal{B}$
        (wasteful tool calls when $\sigma$ is under-used). Explicit
        $\rho$/$\sigma$ governance should reduce variance in
        exploration behavior.
\end{enumerate}

\subsection{Open Problems}
\label{sec:open}

The framework raises several questions that go beyond what the
current formalism can answer:

\begin{itemize}[leftmargin=*]
  \item \textbf{Optimal composition order.} The inbound pipeline
        $\lambda \circ \delta \circ \phi \circ \pi^{+} \circ \sigma$
        prescribes a fixed order. Are there task-dependent orderings
        that improve performance? For instance, should $\delta$
        precede $\phi$ when the simplification itself is
        position-sensitive?
  \item \textbf{Operator interaction effects.} The predictions above
        treat operators as independent. In practice, operators may
        interact: strong $\lambda$ (layering) might partially
        compensate for weak $\delta$ (displacement) by structurally
        isolating high-priority content. Understanding these
        interactions requires factorial experimental designs beyond
        single-operator ablation.
  \item \textbf{Cross-architecture transfer.} The salience geometry
        of \Cref{sec:geometry} is specific to softmax-based
        transformers. State space models and diffusion language models
        (\Cref{sec:nonlinear}) may exhibit different salience
        profiles, potentially rendering $\delta$ unnecessary while
        creating demand for new operators. The completeness argument
        (\Cref{sec:operators}) holds for the current transformer-based
        zonal structure; non-transformer architectures may alter that
        structure and with it the operator basis.
  \item \textbf{Automated operator selection.} Can a meta-controller
        learn which operators to apply and with what parameters, given
        task characteristics and current context state? This connects
        the framework to the broader question of learned context
        management~\citep{packer2024memgpt, memos2025b}.
\end{itemize}

\section{Discussion}
\label{sec:discussion}

\subsection{Which Operators Survive Architecture Change?}
\label{sec:nonlinear}

The seven operators compensate for specific properties of transformer
architectures: linear prefix memory, append-only KV caches, and
entropy growth under softmax. Emerging architectures relax different
subsets of these constraints, and the framework predicts which
operators each architecture renders obsolete versus which it preserves.

State Space Models~\citep{gu2024mamba, dao2024mamba2} and their
production-scale hybrids~\citep{nvidia2025nemotronh, lieber2024jamba,
glorioso2024zamba2}---as well as pure recurrent architectures such as
RWKV-7~\citep{peng2025rwkv7}---compress history into evolving latent
states with constant-memory processing. This eliminates the
append-only constraint, potentially internalizing $\delta$
(displacement): if the model can revise its state representation
in-place, positional salience gradients no longer dictate where
information must be placed. However, the compression is lossy---the
latent state is itself a form of $\pi^{-}$ (compaction) applied
continuously and without external governance.

\begin{conjecture}[Implicit Context Collapse]
\label{conj:icc}
SSM-based agents will exhibit implicit context collapse: information
loss that accumulates inside the latent state without any mechanism to
detect or reverse it---analogous to repeated $\pi^{-}$ without
archival (\Cref{sec:zones}), but executed continuously within the
model's hidden state rather than as a discrete external step.
\end{conjecture}

If \Cref{conj:icc} holds, then $\phi$ (simplification) and
$\sigma_{\textup{expire}}$ (lifecycle-aware selection) become
\emph{more} important under SSMs, not less, because the compaction is
no longer an explicit step that can be audited.

Diffusion language models~\citep{nie2025llada,
arriola2025blockdiffusion, ye2025dream} weaken the left-to-right
constraint, enabling local non-causal refinement. This relaxes the
salience geometry of \Cref{sec:geometry}: if the model can attend
bidirectionally, the U-shaped salience profile flattens, and $\delta$
loses its rationale entirely. But diffusion models introduce a new
constraint---iterative denoising across the full sequence---that
creates computational pressure to keep $|\mathcal{V}|$ small. The
framework predicts that $\phi$ and $\alpha$ (aggregation) will be
critical for diffusion-based agents, not to compensate for attention
non-uniformity but to keep the denoising target tractable.

Architectures that treat context as an external navigable
environment~\citep{zhang2025rlm} or that modify weights at test
time~\citep{behrouz2025titans, zweiger2025seal} push further: they
begin to internalize $\mathcal{G}$ itself, blurring the boundary
between stored memory and model parameters. Under such architectures,
$\pi$ (projection) shifts from an external engineering task to an
internal learned capability. Yet the zonal structure persists: the
world will always exceed any model's capacity ($\mathcal{B} \neq
\emptyset$), memory will always require governance ($\mathcal{G}$
cannot be unbounded), and reasoning will always require a bounded
surface ($|\mathcal{V}|$ is finite). What changes is not the need for
cartographic operators but \emph{where they execute}---externally as
middleware, or internally as learned model behavior.

\subsection{Cartographic Competencies as Training Objectives}
\label{sec:training}

Current LLMs are optimized to continue text, not to manage contextual
space. Yet agent behavior demands cartographic competencies: choosing
between $\rho$ (reconnaissance) and $\sigma$ (selection) when facing
knowledge gaps, applying $\pi$ and $\phi$ to
$\textup{\textsc{recall}}$ outputs, and respecting $\lambda$ constraints
in $\mathcal{V}$. The framework identifies three specific mismatches
between next-token prediction training and cartographic requirements.

First, next-token prediction treats all positions as equally important
prediction targets, but the salience geometry of \Cref{sec:geometry}
shows that not all positions contribute equally to downstream
reasoning. Training paradigms that weight prediction targets by
information density---such as patch-level
training~\citep{shao2025patch} and mutual-information-aware
objectives~\citep{yang2025mutual}---implicitly learn a form of
$\phi$ (simplification): the model learns which tokens carry
structural signal and which are surface syntax.

Second, the $\rho$/$\sigma$ boundary (when to explore $\mathcal{B}$
vs.\ exploit $\mathcal{G}$) is not present in pre-training at all.
Standard training never requires the model to decide ``I don't know
this---I should invoke a tool'' versus ``I saw this earlier---I should
recall it.'' This competency must currently be elicited through
prompting or reinforcement learning on agentic trajectories. The
framework suggests that training explicitly on $\rho$/$\sigma$
decision points---tasks where the optimal action depends on the
agent's epistemic state across zones---would improve tool-use
calibration.

Third, surveys of System~1 to System~2 reasoning in
LLMs~\citep{li2025system2survey, kahneman2011thinking} catalog
methods (MCTS, RL fine-tuning, macro actions) that require the agent
to maintain structured state. \citet{li2025structure} provide direct
evidence that the \emph{structure} of reasoning demonstrations---not
their content---determines reasoning quality: disrupting the logical
organization of chain-of-thought steps collapses accuracy, while
substituting incorrect content has minimal effect. This supports our
argument that deliberate reasoning (System~2) is mediated by how
context is structured: it requires the agent to maintain layered state
($\lambda$), track what has been explored ($\rho$) versus what is
assumed ($\sigma$), and revise beliefs when new evidence arrives.
These are cartographic competencies, and their absence under standard
training explains why scaling model size alone does not reliably
produce agentic capability.

\subsection{Multi-Agent Cartography}
\label{sec:multiagent}

Multi-agent systems~\citep{tran2025multiagent} extend the framework
from a single zonal structure to a \emph{federation of maps}. Each
agent $i$ maintains its own $(\mathcal{B}_i, \mathcal{G}_i,
\mathcal{V}_i)$, and agents may share a common memory
$\mathcal{G}_{\textup{shared}}$ with layered provenance ($\lambda$
applied across agent boundaries). Cross-agent communication is modeled
as constrained $\pi^{+}$: agent $i$ projects a result from
$\mathcal{V}_i$ into $\mathcal{G}_{\textup{shared}}$ (or directly
into $\mathcal{V}_j$) using a projection schema that both parties
must agree on. The central question is how these maps interact.

The framework identifies three distinct multi-agent failure modes, each
corresponding to a specific operator failure across agent boundaries:

\begin{itemize}[leftmargin=*, nosep]
  \item \textbf{Projection incompatibility.} Agent $A$ applies
        $\pi^{+}$ with schema $p_A$ to communicate a finding; agent $B$
        receives it but interprets it under schema $p_B$. If
        $p_A \neq p_B$, the receiving agent's $\mathcal{V}$ contains
        well-formed but semantically distorted content---the
        cross-agent projection incoherence identified in
        \Cref{sec:zones}.
  \item \textbf{Namespace collision.} Two agents write to a shared
        $\mathcal{G}$ without coordinated $\lambda$ (layering). Their
        contributions are interleaved without provenance marking,
        making it impossible for a third agent to distinguish or
        prioritize between them. This is the multi-agent analogue of
        the single-agent layering failure.
  \item \textbf{Reconnaissance duplication.} Multiple agents
        independently apply $\rho$ to the same region of $\mathcal{B}$,
        wasting exploration budget. Without a shared record of what
        has been explored, the system cannot distinguish
        $\mathcal{B}$ (unexplored) from $\mathcal{G}$ (explored by
        another agent but not yet shared).
\end{itemize}

Global Workspace Theory~\citep{baars1988cognitive, goldstein2024gwt}
offers an architectural response: a governed broadcast
bottleneck---analogous to a shared $\mathcal{V}$---through which all
inter-agent communication must pass. In cartographic terms, this
shared workspace forces agents to use a \emph{common projection
schema} for cross-agent $\pi^{+}$, a \emph{coordinated namespace} for
shared $\mathcal{G}$ via $\lambda$, and a \emph{joint exploration
ledger} for $\rho$. Empirical evidence supports this:
\citet{yuen2025intrinsic} demonstrate that agent-specific structured
memories shared across heterogeneous agents achieve state-of-the-art
coordination, and cognitive-workspace
frameworks~\citep{an2025workspace, li2025sculptor} show that
structured broadcast improves coherence even without specialized
training---consistent with the prediction that the benefit comes from
operator coordination rather than model capability.

\section{Limitations}
\label{sec:limitations}

Several limitations of this work should be acknowledged. First, the
framework is primarily \emph{interpretive}: the cartographic operators
are derived from cartographic generalization theory, and while we
distinguish structural isomorphisms from functional analogies
(\Cref{sec:operators}), the formalization does not yet constitute a
full mathematical theory. Second, the case study analysis
(\Cref{sec:practices}) is post-hoc---the operators were informed in
part by observing these systems, so the convergence finding is
interpretive rather than predictive. Third, while the operators are
derived from systematic coverage of zone transformations, additional
cartographic operators (e.g., symbolization, rotation) may have context
analogues not yet identified, and the zone-general scope of
$\phi$, $\alpha$, and $\lambda$ may warrant finer distinctions as
empirical evidence accumulates. Fourth, the research agenda (\Cref{sec:agenda}) derives testable
predictions but does not empirically validate them; the proposed CCD
benchmark requires community implementation and evaluation.
Fifth, the zonal model is defined for a single agent; the multi-agent
extension (\Cref{sec:multiagent}) identifies failure modes and
architectural responses but does not formally extend the zone
definitions or operator signatures to federated settings.
Finally, the extension of cartographic theory from geographic to
contextual space introduces representational differences: contextual
space is sequential and high-dimensional, while geographic space is
continuous and low-dimensional. The extent to which cartographic
principles transfer across this gap---and whether the invariant triad
(semantic, structural, salience) fully characterizes the contextual
distortion space---requires further investigation.

\section{Conclusion}
\label{sec:conclusion}

We have presented Context Cartography, a formal framework for
governing contextual space in LLM systems. The framework contributes a
tripartite zonal model with explicit state transitions, a
characterization of salience geometry grounding governance strategies
in attention physics, and seven formally defined cartographic
operators---reconnaissance, selection, simplification, aggregation,
projection, displacement, and layering---derived from systematic
coverage of all zone transformations and organized by transformation
type and zone scope. Analysis of four contemporary systems provides
interpretive evidence that these operators are converging independently
across the industry.

The central claim of this work is that contextual space has geometry,
and that intelligence depends on how that geometry is governed. Current
cartographic practices are compensatory---they exist because no
architecture natively supports the full range of spatial governance
that agentic reasoning demands. As architectures evolve, some
operators may internalize (transitioning from external engineering to
learned model behavior), but the zonal structure persists: the world
exceeds any model's capacity, memory requires governance, and
reasoning requires a bounded surface. The testable predictions and
diagnostic benchmark proposed in our research agenda provide a path
toward empirical validation.

\bibliographystyle{plainnat}
\bibliography{references}

\clearpage
\appendix

\section{Operator Implementation Scoring Evidence}
\label{app:scoring}

This appendix provides the evidence basis for each cell in
\Cref{tab:convergence}. Each system--operator pair is evaluated on
five binary criteria: \textbf{P}resent (any mechanism performs this
transformation), \textbf{E}xplicit (named architectural component),
\textbf{C}onfigurable (parameters tunable by user/developer),
\textbf{A}utomated (occurs without manual intervention), and
\textbf{D}ocumented (described in official documentation or papers).
Scores range from 0 (absent) to 5 (all criteria met).

\subsection{Claude Code}
\label{app:claude}

\begin{description}[leftmargin=0pt, labelindent=0pt, style=nextline]

\item[$\rho$ (Reconnaissance) = 5 \textnormal{(P,E,C,A,D)}]
Subagent spawning via the Task tool is the core exploration
mechanism~\citep{anthropic2025claudecode}. \textbf{P}: Subagents
traverse $\mathcal{B}$ by executing file searches, code reads, and
web fetches in isolated contexts. \textbf{E}: Named as ``Task tool''
with typed agent variants (Explore, Plan, general-purpose).
\textbf{C}: Users specify agent type, prompt, and isolation mode
(e.g., worktree). \textbf{A}: The orchestrator autonomously decides
when to spawn subagents. \textbf{D}: Documented in Claude Code
official documentation.

\item[$\sigma$ (Selection) = 2 \textnormal{(P,A)}]
\textbf{P}: Subagents implicitly filter what to return to the
orchestrator---not all explored content is surfaced. \textbf{A}: This
filtering is automated within the subagent's reasoning. Not
\textbf{E}: no named ``selection'' component exists. Not \textbf{C}:
no user-facing parameter controls filtering granularity. Not
\textbf{D}: not described as a distinct feature.

\item[$\phi$ (Simplification) = 4 \textnormal{(P,E,A,D)}]
\textbf{P}: Subagent results are compressed before return to the
orchestrator. \textbf{E}: Described as ``condensed results'' in
documentation. \textbf{A}: Compression is automatic---subagents
return summaries, not raw tool output. \textbf{D}: Documented as part
of the subagent architecture. Not \textbf{C}: no user-configurable
compression level or summary length parameter.

\item[$\alpha$ (Aggregation) = 1 \textnormal{(P)}]
\textbf{P}: When multiple subagents return results, their outputs
coexist in the orchestrator's context, but there is no explicit
mechanism to fuse or deduplicate them. Not \textbf{E,C,A,D}.

\item[$\pi$ (Projection) = 2 \textnormal{(P,A)}]
\textbf{P}: Subagent results are projected from the subagent's
internal representation into the orchestrator's token sequence.
\textbf{A}: This happens automatically upon subagent completion. Not
\textbf{E}: no named projection schema. Not \textbf{C}: results are
returned at a single fixed resolution---no multi-resolution control.
Not \textbf{D}: not described as a distinct projection mechanism.

\item[$\delta$ (Displacement) = 2 \textnormal{(P,A)}]
\textbf{P}: Subagent summaries enter the orchestrator's context at
the recency-biased end of the sequence, implicitly placing them in a
high-salience position. \textbf{A}: This is an automatic consequence
of append-only context. Not \textbf{E}: no named reordering
mechanism. Not \textbf{C}: position is determined by insertion order,
not configurable. Not \textbf{D}: not described as positional
management.

\item[$\lambda$ (Layering) = 5 \textnormal{(P,E,C,A,D)}]
\textbf{P}: System prompts, user messages, tool results, and
CLAUDE.md instructions occupy distinct typed namespaces.
\textbf{E}: Named layers include ``system-reminder,'' ``tool
results,'' and ``user messages.'' \textbf{C}: Users configure layers
via CLAUDE.md, system prompts, and hook configurations. \textbf{A}:
Layer separation is enforced automatically by the runtime.
\textbf{D}: Documented in Claude Code documentation.

\end{description}

\subsection{Letta (MemGPT)}
\label{app:letta}

\begin{description}[leftmargin=0pt, labelindent=0pt, style=nextline]

\item[$\rho$ (Reconnaissance) = 1 \textnormal{(P)}]
\textbf{P}: Letta agents can invoke tools and external APIs. Not
\textbf{E,C,A,D}: tool use is a general capability, not an
architecturally distinct exploration mechanism.

\item[$\sigma$ (Selection) = 2 \textnormal{(P,A)}]
\textbf{P}: Recall memory search retrieves relevant archived content.
\textbf{A}: Search is triggered automatically during context assembly.
Not \textbf{E}: selection is embedded within the memory hierarchy, not
a standalone component. Not \textbf{C,D}.

\item[$\phi$ (Simplification) = 4 \textnormal{(P,E,A,D)}]
\textbf{P}: Evicted content undergoes recursive
summarization~\citep{packer2024memgpt}. \textbf{E}: Named as
``recursive summarization'' in the MemGPT paper. \textbf{A}: Triggered
automatically when context overflows. \textbf{D}: Described in the
MemGPT paper and Letta documentation~\citep{letta2025context}. Not
\textbf{C}: summarization parameters are not user-configurable.

\item[$\alpha$ (Aggregation) = 2 \textnormal{(P,A)}]
\textbf{P}: Sleep-time background processes reorganize and merge
related memory entries. \textbf{A}: Runs asynchronously without user
intervention. Not \textbf{E}: not a named standalone component. Not
\textbf{C,D}.

\item[$\pi$ (Projection) = 5 \textnormal{(P,E,C,A,D)}]
\textbf{P}: Core/recall memory tiers implement multi-resolution
projection~\citep{packer2024memgpt}. \textbf{E}: Named as ``core
memory'' (full resolution) and ``recall memory'' (compressed
resolution). \textbf{C}: Users configure core memory block contents
and structure. \textbf{A}: Projection from recall to context is
automatic. \textbf{D}: Central architectural contribution of the
MemGPT paper.

\item[$\delta$ (Displacement) = 1 \textnormal{(P)}]
\textbf{P}: Content position is determined by insertion order and
memory tier, providing implicit positional bias. Not \textbf{E,C,A,D}:
no explicit positional management mechanism.

\item[$\lambda$ (Layering) = 5 \textnormal{(P,E,C,A,D)}]
\textbf{P}: Core memory blocks (human, persona, system) are typed
namespaces~\citep{packer2024memgpt}. \textbf{E}: Named block types
with distinct roles. \textbf{C}: Users define and modify block
contents. \textbf{A}: Block separation is maintained automatically.
\textbf{D}: Documented in MemGPT paper and Letta platform.

\end{description}

\subsection{MemOS}
\label{app:memos}

\begin{description}[leftmargin=0pt, labelindent=0pt, style=nextline]

\item[$\rho$ (Reconnaissance) = 1 \textnormal{(P)}]
\textbf{P}: Agents can query external sources. Not \textbf{E,C,A,D}:
general capability, not an architecturally distinct exploration
mechanism.

\item[$\sigma$ (Selection) = 5 \textnormal{(P,E,C,A,D)}]
\textbf{P}: DAG-based scheduling selects which memory nodes to
activate~\citep{memos2025, memos2025b}. \textbf{E}: Named as
``MemScheduler'' with DAG-based dependency resolution. \textbf{C}:
Users configure scheduling policies and memory access patterns.
\textbf{A}: Next-scene prediction preloads memory before explicit
request. \textbf{D}: Described in both MemOS papers.

\item[$\phi$ (Simplification) = 2 \textnormal{(P,A)}]
\textbf{P}: Graph operations may reduce information during traversal.
\textbf{A}: Occurs during automated graph queries. Not \textbf{E}:
simplification is not a named standalone mechanism. Not \textbf{C,D}.

\item[$\alpha$ (Aggregation) = 4 \textnormal{(P,E,A,D)}]
\textbf{P}: Graph-structured memory aggregates related entities into
composite nodes~\citep{memos2025}. \textbf{E}: MemCube merges are
explicit graph operations. \textbf{A}: Aggregation occurs during graph
maintenance. \textbf{D}: Described in MemOS papers. Not \textbf{C}:
aggregation policies are not directly user-configurable.

\item[$\pi$ (Projection) = 5 \textnormal{(P,E,C,A,D)}]
\textbf{P}: Graph $\to$ text projection with multiple
schemas~\citep{memos2025b}. \textbf{E}: Named projection modes
include entity lists, relationship summaries, and subgraph traversals.
\textbf{C}: Projection schema is selectable per query. \textbf{A}:
Projection is automatic during memory retrieval. \textbf{D}: Described
in both MemOS papers.

\item[$\delta$ (Displacement) = 1 \textnormal{(P)}]
\textbf{P}: No explicit positional management; content order is
determined by graph traversal order. Not \textbf{E,C,A,D}.

\item[$\lambda$ (Layering) = 5 \textnormal{(P,E,C,A,D)}]
\textbf{P}: MemCube abstraction isolates knowledge bases as typed
containers~\citep{memos2025}. \textbf{E}: Named as ``MemCube'' with
composable isolation. \textbf{C}: Users create and configure distinct
MemCubes. \textbf{A}: Isolation is maintained automatically.
\textbf{D}: Central contribution of the MemOS architecture.

\end{description}

\subsection{OpenViking}
\label{app:openviking}

\begin{description}[leftmargin=0pt, labelindent=0pt, style=nextline]

\item[$\rho$ (Reconnaissance) = 1 \textnormal{(P)}]
\textbf{P}: Agents can traverse the filesystem hierarchy to discover
content. Not \textbf{E,C,A,D}: traversal is a general retrieval
mechanism, not a named exploration component.

\item[$\sigma$ (Selection) = 5 \textnormal{(P,E,C,A,D)}]
\textbf{P}: Path-based directory traversal selects content by
hierarchical location~\citep{openviking2025}. \textbf{E}: Named as
structured traversal with custom URI scheme. \textbf{C}: Users
configure directory structure and retrieval paths. \textbf{A}:
Selection follows directory recursion automatically. \textbf{D}:
Documented in OpenViking repository.

\item[$\phi$ (Simplification) = 1 \textnormal{(P)}]
\textbf{P}: L0 summaries (${\sim}$100 tokens) are brief, but this is
a property of the projection tier, not a standalone simplification
mechanism. Not \textbf{E,C,A,D} as a distinct operator.

\item[$\alpha$ (Aggregation) = 1 \textnormal{(P)}]
\textbf{P}: No explicit aggregation mechanism; retrieved content is
presented individually. Not \textbf{E,C,A,D}.

\item[$\pi$ (Projection) = 5 \textnormal{(P,E,C,A,D)}]
\textbf{P}: L0/L1/L2 tiered loading implements multi-resolution
projection~\citep{openviking2025}. \textbf{E}: Named tiers with
explicit token budgets (L0: ${\sim}$100, L1: ${\sim}$1{,}000, L2:
full). \textbf{C}: Tier selection is configurable per retrieval
request. \textbf{A}: Tier assignment is automatic based on query
context. \textbf{D}: Documented as the core architecture.

\item[$\delta$ (Displacement) = 1 \textnormal{(P)}]
\textbf{P}: No explicit positional management; retrieval order
follows filesystem hierarchy. Not \textbf{E,C,A,D}.

\item[$\lambda$ (Layering) = 5 \textnormal{(P,E,C,A,D)}]
\textbf{P}: Hierarchical filesystem with custom URI scheme enforces
namespace separation~\citep{openviking2025}. \textbf{E}: Named
directory hierarchy with typed content. \textbf{C}: Users configure
the filesystem structure. \textbf{A}: Namespace isolation is
maintained by the filesystem abstraction. \textbf{D}: Documented in
OpenViking repository.

\end{description}

\section{Lean 4 Formalization}
\label{app:lean}

Five structural properties of the framework have been formalized and
mechanically verified in Lean~4 using Mathlib. All proofs compile
with only standard axioms (\texttt{propext}, \texttt{Classical.choice},
\texttt{Quot.sound}). Theorem statements are shown below; full proofs are available at
\url{https://github.com/lucifer1004/context-cartography}.

\subsection{Unique Zone Membership (Definition~1)}

Every element belongs to exactly one zone, derived from the
disjointness and coverage axioms of the \texttt{ContextState}
structure.

\begin{lstlisting}
theorem mem_unique_zone (s : ContextState U) (x : U) :
    (x @$\in$@ s.blackFog @$\wedge$@ x @$\notin$@ s.grayFog @$\wedge$@ x @$\notin$@ s.visible) @$\vee$@
    (x @$\notin$@ s.blackFog @$\wedge$@ x @$\in$@ s.grayFog @$\wedge$@ x @$\notin$@ s.visible) @$\vee$@
    (x @$\notin$@ s.blackFog @$\wedge$@ x @$\notin$@ s.grayFog @$\wedge$@ x @$\in$@ s.visible)
\end{lstlisting}

\subsection{Boundary Operator Uniqueness (\S\ref{sec:composition})}

Each boundary transition is governed by exactly one operator kind.
Removing any boundary operator leaves its transition without
representational governance.

\begin{lstlisting}
-- @$\rho$@ is the sole operator for B @$\to$@ G
theorem reconnaissance_sole :
    @$\forall$@ op, operatorCovers op (blackFog, grayFog) @$\to$@
    op = .reconnaissance

-- @$\sigma$@ is the sole operator kind for G @$\to$@ B
theorem selection_sole :
    @$\forall$@ op, operatorCovers op (grayFog, blackFog) @$\to$@
    @$\exists$@ m, op = .selection m

-- @$\pi^{+}$@ is the sole non-selection operator for G @$\to$@ V
theorem forwardProjection_sole :
    @$\forall$@ op, operatorCovers op (grayFog, visible) @$\to$@
    op = .forwardProjection @$\vee$@ @$\exists$@ m, op = .selection m

-- @$\pi^{-}$@ is the sole non-selection operator for V @$\to$@ G
theorem inverseProjection_sole :
    @$\forall$@ op, operatorCovers op (visible, grayFog) @$\to$@
    op = .inverseProjection @$\vee$@ @$\exists$@ m, op = .selection m
\end{lstlisting}

\subsection{Composition Non-Commutativity (\S\ref{sec:composition})}

A concrete witness over \texttt{Fin}\,\texttt{3} proving that
$\delta \circ \phi \neq \phi \circ \delta$: $\phi$ keeps only element~0,
$\delta$ adds element~1.
Then $\delta(\phi(\{0,2\})) = \{0,1\}$ but
$\phi(\delta(\{0,2\})) = \{0\}$.

\begin{lstlisting}
theorem composition_noncommutative :
    @$\delta_1$@(@$\varphi_0$@.simplify {0, 2}) @$\neq$@ @$\varphi_0$@.simplify (@$\delta_1$@ {0, 2})
\end{lstlisting}

\subsection{Context Collapse (\S\ref{sec:zones})}

Destructive compaction (without archival) provably loses information
to $\mathcal{B}$; archival compaction preserves all content in
$\mathcal{G} \cup \mathcal{V}$.

\begin{lstlisting}
-- Destructive: non-summary V content moves to B
theorem destructive_loses_to_blackfog
    (s : ContextState U) (summary : Set U)
    (x : U) (hx_vis : x @$\in$@ s.visible)
    (hx_not : x @$\notin$@ summary) :
    x @$\in$@ (destructiveCompaction s summary).newBlackFog

-- Archival: all original V content remains in G @$\cup$@ V
theorem archival_preserves_all
    (s : ContextState U) (summary : Set U) :
    s.visible @$\subseteq$@
      (archivalCompaction s summary).newGrayFog @$\cup$@
      (archivalCompaction s summary).newVisible
\end{lstlisting}

\subsection{Invariant Impossibility (\S\ref{sec:properties})}

No strict reduction can simultaneously preserve semantic content
(critical elements retained), structural integrity (linked elements
co-occur), and salience priority (high-priority element retained).
Over \texttt{Fin}\,\texttt{3}, element~0 is semantically critical, element~0
structurally requires element~1, and element~2 is high-priority.
Any strict reduction from $\{0,1,2\}$ must drop at least one,
violating at least one invariant.

\begin{lstlisting}
theorem invariant_impossibility :
    @$\neg$@ @$\exists$@ (f : Set (Fin 3) @$\to$@ Set (Fin 3)),
      f {0, 1, 2} @$\subset$@ {0, 1, 2} @$\wedge$@
      preservesSemantic f (@$\cdot$@ = 0) {0, 1, 2} @$\wedge$@
      preservesStructural f
        (fun x y @$\Rightarrow$@ x = 0 @$\wedge$@ y = 1) {0, 1, 2} @$\wedge$@
      preservesSalience f 2 {0, 1, 2}
\end{lstlisting}

\noindent The proof proceeds by deriving $0 \in f(S)$ from semantic
preservation, $1 \in f(S)$ from structural preservation, and
$2 \in f(S)$ from salience preservation, yielding
$f(S) \supseteq S$---contradicting strict reduction.

\end{document}